\pdfoutput=1

\documentclass[11pt]{article}

\usepackage[hyphens]{url}
\usepackage[]{acl}

\usepackage{times}
\usepackage{latexsym}

\usepackage[T1]{fontenc}

\usepackage[utf8]{inputenc}

\usepackage{microtype}

\newif\ifarxiv
\usepackage{sec/package}

\usepackage{tikz}
\usetikzlibrary{decorations.pathreplacing}
\usepackage{subcaption}
\usepackage{graphicx}

\usepackage{booktabs}
\usepackage{multirow}

\usepackage{amsmath}
\usepackage{amsfonts}
\usepackage{amsthm}
\newtheorem{definition}{Definition}

\usepackage{csquotes}

\DeclareMathOperator*{\supp}{supp}

\newcommand{\xx}{\boldsymbol{x}}

\newcommand*{\X}{\mathcal{X}}

\newcommand*{\M}{\mathcal{M}}

\newcommand*{\pr}[1]{\mathrm{Pr}\left[#1\right]}

\everypar{\looseness=-1}

\usepackage{xcolor}
\definecolor{color1}{RGB}{102,194,165}
\definecolor{color2}{RGB}{141,160,203}
\definecolor{color3}{RGB}{252,141,98}
\definecolor{light-gray}{gray}{0.8}

\usepackage{cleveref}
\crefname{section}{\S}{\S\S}
\Crefname{section}{\S}{\S\S}
\crefname{table}{Tab.}{}
\crefname{figure}{Fig.}{}
\crefname{algorithm}{Alg.}{}
\crefname{appendix}{App.}{}
\crefname{lemma}{Lemma}{}
\Crefname{theorem}{Theorem}{}
\crefname{prop}{Proposition}{}
\crefname{cor}{Corollary}{}
\crefname{align}{}{}
\crefname{equation}{}{}

\usepackage{todonotes}
\usepackage{inconsolata}

\title{
Differentially Private Language Models for Secure Data Sharing
}

\author{Justus Mattern \\
  RWTH Aachen \\
  \texttt{justus.mattern@rwth-aachen.de} \\\And
  Zhijing Jin \\
  MPI \& ETH Zürich \\
  \texttt{zjin@tue.mpg.de} \\\And
  Benjamin Weggenmann \\
  SAP Security Research \\
  \texttt{benjamin.weggenmann@sap.com} \\\AND
  Bernhard Schölkopf\thanks{\hspace{0.1cm} Equal Supervision.} \\
  MPI for Intelligent Systems \\
  \texttt{bs@tue.mpg.de} \\\And
    Mrinmaya Sachan\samethanks{} \\
  ETH Zürich \\
  \texttt{msachan@ethz.ch}\\}

\begin{document}
\maketitle
\begin{abstract}
To protect the privacy of individuals whose data is being shared, it is of high importance to develop methods allowing researchers and companies to release textual data while providing formal privacy guarantees to its originators. In the field of NLP, substantial efforts have been directed at building mechanisms following the framework of local differential privacy, thereby anonymizing individual text samples before releasing them. In practice, these approaches are often dissatisfying in terms of the quality of their output language due to the strong noise required for local differential privacy. In this paper, we approach the problem at hand using global differential privacy, particularly by training a generative language model in a differentially private manner and consequently sampling data from it. Using natural language prompts and a new prompt-mismatch loss, we are able to create highly accurate and fluent textual datasets taking on specific desired attributes such as sentiment or topic and resembling statistical properties of the training data. We perform thorough experiments indicating that our synthetic datasets do not leak information from our original data and are of high language quality and highly suitable for training models for further analysis on real-world data. Notably, we also demonstrate that training classifiers on private synthetic data outperforms directly training classifiers on real data with DP-SGD.\footnote{Our code is available at \url{https://github.com/justusmattern/private-datasets-with-llms}.
}
\end{abstract}

\section{Introduction}
Rapid advancements in the field of deep learning and natural language processing (NLP) have enabled companies, public institutions and researchers to extract information and gain knowledge from large-scale data generated by individuals. In many cases, it is desirable to share such data with third parties, for example when analyses are performed by external consultants or in order to provide high quality benchmarks for the research community. This, however, entails a variety of risks related to privacy that cannot merely be solved by pseudonymization: A variety of deanonymization attacks enable the re-identification of individuals from tabular data such as movie ratings \citep{netflix-narayanan}, geolocation data \citep{deanonymization-social-lee} and notably also text \citep{koppel2009computational, shrestha-etal-2017-convolutional, fabien-etal-2020-bertaa}. It is therefore highly desirable to develop anonymization mechanisms enabling secure data sharing, ideally with mathematical privacy guarantees as granted by differential privacy (DP) \citep{dwork2014algorithmic}.


\begin{figure}[t]
    \centering
    \includegraphics[width=1.0\linewidth]{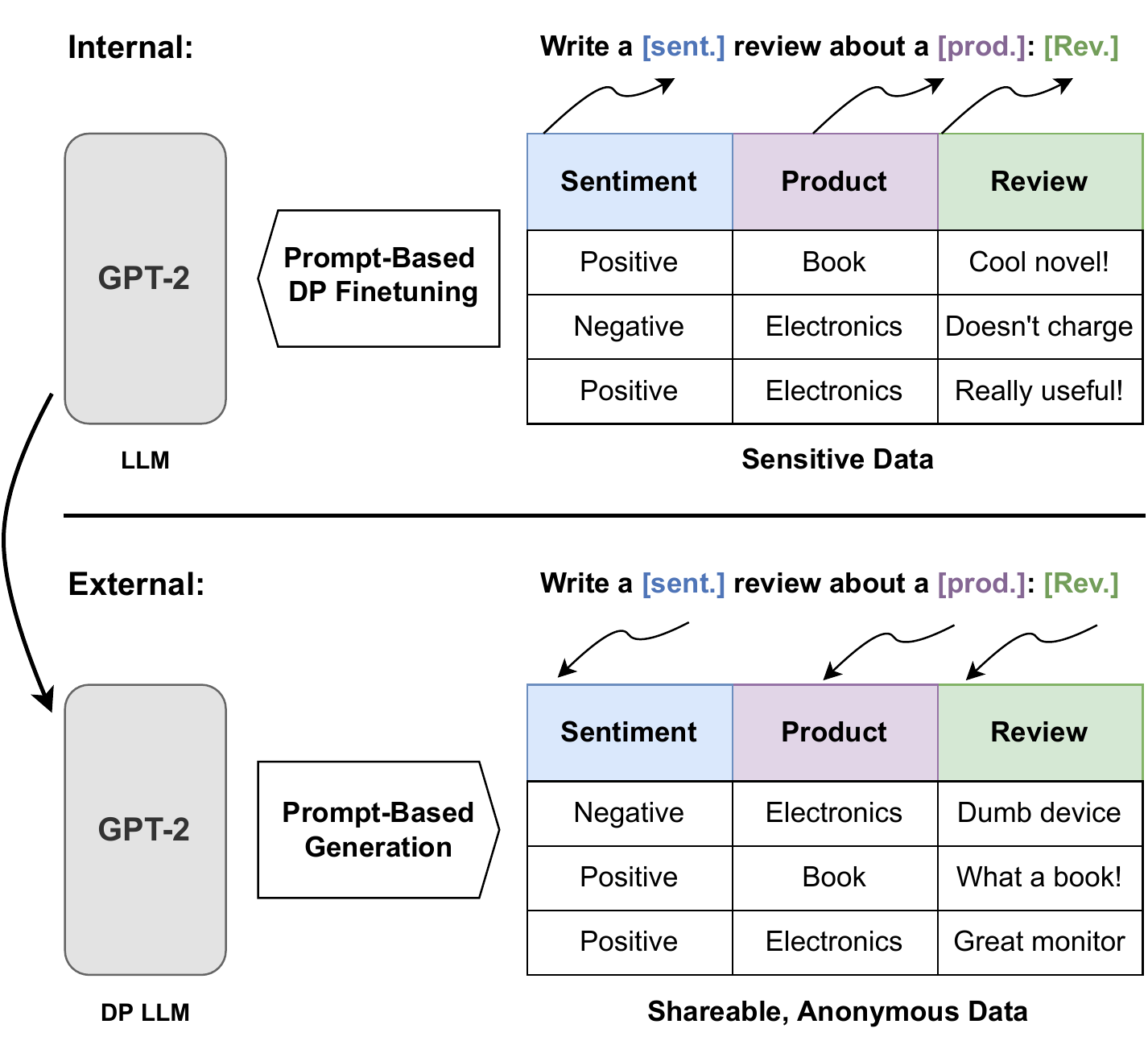}
    \caption{Main idea of our paper: To share potentially sensitive datasets with third parties, we train a language model (LM) on the sensitive data in a differentially private manner and consequently prompt the LM to generate synthetic samples with privacy guarantees.}
    \label{fig:overview}
\end{figure}

Existing approaches anonymize every text sample individually by obtaining differentially private vector representations \citep{syntf-weggenmann, fernandes2019generalised} or using sequence-to-sequence approaches that rewrite a given sample to eliminate user-revealing information \citep{a4nt-shetty, feyisetan-leveraging, feyisetan-textual,  dp-vae-weggenmann}, thereby following local differential privacy. As pointed out by \citet{mattern-limits}, local DP requires a very high degree of noise which often leads to incoherent language and only little semantic overlap. The strict requirements of local DP are, however, not necessary if we assume that an entity aiming to share data already has access to the full collection of user-written texts and only wants to release an anonymized version of it.

In this paper, inspired by recent advances demonstrating the feasibility of training large language models (LLMs) in a differentially private manner \citep{li2021large}, we propose a globally differentially private data release mechanism relying on the generation of a "twin" dataset of the original, sensitive user data from large language models.
As depicted in Figure \ref{fig:overview}, we train GPT-2 \citep{radford2019language} to generate texts of our original dataset based on prompts inferred from the sample's individual attributes such as sentiment or topic. For fine-tuning, we use a differentially private optimization algorithm in order to protect the content of our training data. Subsequently, we sample from the trained model to generate a large number of synthetic, anonymous texts, resulting in a verifiably private "twin" dataset. 
We carefully evaluate our proposed method using popular NLP datasets such as IMDb movie reviews or Amazon product reviews. Here, we find that even after learning with strong privacy guarantees such as $\epsilon = 3$ or $\epsilon = 8$ from only a very limited amount of training samples such as 25 or 50, our generated data is of high quality and the classifiers trained on it achieve accuracies only $\sim$3\% lower than those trained on the full original dataset containing thousands of samples. Notably, we also find that transformer based classification models trained on private data outperform models trained on real data with differentially private optimization. Finally, we show that the differentially private fine-tuning procedure effectively minimizes the risk of data leakage from language models that was previously discovered by \citet{lm-extractdata}.

\section{Background}
\subsection{Differential Privacy}

Differential privacy (DP) is a formal notion of privacy
that is currently considered the state-of-the-art
for quantifying and limiting information disclosure about individuals.
It has been introduced by \citet{dwork2006calibrating}
under the name \emph{$\epsilon$-indistinguishability} with the goal
of giving semantic privacy by quantifying the risk of an individual
that results from participation in data collection.

In the original, \emph{central model} of DP,
we consider \emph{adjacent} datasets that differ by at most one record
(i.e., one individual's data).
A differentially private query on both databases should yield matching results
with similar probabilities,
i.e., answers that are probabilistically \emph{indistinguishable}.
This is achieved via random mechanisms that return noisy query results,
thus masking the impact of each individual.

\begin{definition}
	\label{def:diffpriv}
	Let $\epsilon > 0$ be a privacy parameter, and $0 \leq \delta \leq 1$.
	A randomized mechanism $\M$ on $\X$ fulfills \emph{$(\epsilon,\delta)$-DP}
	if for any pair of adjacent inputs $\xx,\xx' \in \X$,
	and all sets of possible outputs $Z \subset \supp \M$,
	\begin{align}
	    \pr{ \M(\xx) \in Z } \leq e^\epsilon \cdot \pr{ \M(\xx') \in Z } + \delta
	    ~.
	\end{align}
\end{definition}

In the \emph{local model} \citep{duchi2013local},
noise is added locally at the data source, before the data is collected
and stored in a central database.
A basic example is randomized response \citep{warner1965randomized},
where each survey participant either provides a truthful or a random answer
depending on the flip of an (unbiased) coin.
The local model makes the strong assumption that any two inputs are considered adjacent,
which often makes it difficult to achieve a satisfying privacy-utility trade-off.

\subsection{Differentially Private Optimization}
An important application of DP is privacy-preserving machine learning
to protect the privacy of the training data.
Typically, neural networks are trained by optimizing a loss function
using stochastic gradient descent (SGD) or a derived method such as Adam \citep{kingma2014adam},
which iteratively compute gradients of the loss function over batches
of samples from the training dataset.
As shown by \citet{song2013stochastic,bassily2014private,abadi2016learning},
it is possible to implement a differentially private version of SGD (DP-SGD)
by clipping the gradients and applying the Gaussian mechanism \cite{dwork2014algorithmic}: 
The latter works by applying noise from an isotropic Gaussian distribution
$\mathcal{N}(\mathbf{0},\sigma^2\mathbf{I})$, where the standard deviation $\sigma$
is derived based on the desired privacy parameters $\epsilon$ and $\delta$.

To achieve good privacy-utility trade-offs, 
it is important to accurately track the total privacy budget spent
throughout the entire training.
In the context of DP, repeated executions of the same (here: Gaussian) mechanism
is referred to as \emph{composition}. Basic \cite{dwork2006our}
and various more refined, advanced \emph{composition theorems}
\cite{dwork2010boosting,dwork2016concentrated,bun2016concentrated}
have been stated in the literature that aim at providing tight bounds
for the overall privacy budget.
However, these advances still resulted in relatively loose bounds
and thus large overall privacy budgets over the course of highly iterative
algorithms such as DP-SGD.
Tight worst-case bounds for composition were derived by \citet{pmlr-v37-kairouz15},
however, it was shown to be computationally infeasible to compute them
in general \cite{murtagh2016complexity}.

For this reason, specific efforts have been made to find tighter bounds
and accurate approximations for the overall privacy loss:
A first example that provides substantial reduced upper bounds
is the moments accountant \citep{abadi2016learning},
which is closely related to Rényi DP \citep{mironov2017renyi},
a generalization of DP based on Rényi divergence.
Gaussian and $f$-DP \cite{dong2019gaussian} provide an approximation
of the total budget using the central limit theorem (CLT).
Finally, \citet{gopi2021numerical,koskela2020computing}, inspired by \citet{sommer2019privacy},
are able to compute the exact budget numerically up to arbitrary precision by aggregating
the \emph{privacy loss random variable} with fast Fourier transform.

\section{Approach}

We consider the following scenario to motivate our approach: an entity wants to implement NLP pipelines to gain insights from internal data, e.g., emails from customers. To seek advice and get support for modeling the data and building pipelines, the entity aims to share an excerpt of the internal data with a third party such as a consultant or a group of researchers. In order to do this without compromising the privacy of its customers, the aim is to synthesize a verifiably private ``toy'' dataset that reflects the properties of the original data without leaking private information. On such a toy dataset, a third party could research how to best solve the task at hand and train a model to perform inference on the actual internal data, without being able to access sensitive information about customers.
Formally, we aim to achieve the following goal: We consider a dataset 
consisting of a training set $\mathcal{D}_{\mathrm{train}}$ and test set $\mathcal{D}_{\mathrm{test}}$. Given \(\mathcal{D}_{\mathrm{train}}\) or 
a subset of it,
we want to train a generative model to synthesize a dataset \(\widetilde{\mathcal{D}}_{\mathrm{train}}\) that does not leak information from the original \(\mathcal{D}_{\mathrm{train}}\). Furthermore, the synthesized dataset should share statistical properties with the original one so that a classification model 
trained on \(\widetilde{\mathcal{D}}_{\mathrm{train}}\) performs as well as if it was trained on \(\mathcal{D}_{\mathrm{train}}\) when making predictions about \(\mathcal{D}_{\mathrm{test}}\).

To achieve this, we use the pretrained autoregressive transformer model \citep{NIPS2017_attention} GPT-2 \citep{radford2019language} and use natural language prompts to enable the conditional generation of text based on desired textual attributes such as its sentiment, domain or genre provided in the prompt. Furthermore, we introduce a new training objective that penalizes the generation of samples fitting another label to reduce the risk of faulty labeled samples in our synthetic dataset.
Finally, we fine-tune our model using a differentially private optimizer to provide privacy guarantees for our training data and to prevent information leakage from our model when subsequently sampling our synthetic dataset.


\subsection{Conditional text generation with natural language prompts}

As we want to control specific textual attributes of our synthetic data, we need to train our model in a manner that allows us to generate different types of texts corresponding to the desired attributes or labels present in our dataset.
We consider a text sample to correspond to a set of $M$ attributes of interest, namely $A := \{a_1, a_2, \dots, a_M\}$, where each attribute $a_j$ can take on a set of categorical values $C_j$.
In the case of product reviews, $a_1$ could be the sentiment of a review that can take on the values $a_1\in C_1 = \{\mathrm{Positive}, \mathrm{Negative}\}$ and $a_2$ can be the product category, so that $a_2 \in C_2=\{\mathrm{Books}, \mathrm{Electronics}, \mathrm{DVD}, \mathrm{Kitchen}\}$. Our goal is to learn a model $p(x|a_1, ...,a_M)$ in order to controllably synthesize text samples according to our desired attributes. 

\begin{figure}[t]
    \centering
    \includegraphics[width=1.0\linewidth]{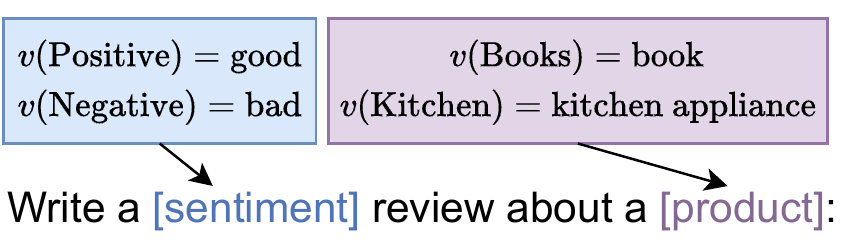}
    \caption{Our template-based approach for generating task instructions. A template consists of placeholders for verbalizations of different attribute values.}
    \label{fig:prompts}
\end{figure}

A straightforward approach to realize this would be to train a single generative model for all possible attribute value combinations. This approach is, however, highly memory-intensive, as it requires us to store the weights of a large number of models that grows exponentially with the number of categorical attributes. Following recent work \citep{schick-schutze-2021-shot}, we therefore train a single language model to conditionally generate texts based on task instructions. Beyond reducing our memory needs, this approach allows us to leverage our model's pretraining knowledge and to perform text generation with only very little training samples \citep{schick-schutze-2021-shot}. Our instructions $\bm{i}(a_1, .., a_M)$ are formed using a template with placeholders that are filled out with verbalizations $v(a_j)$ taking on different forms for different values of every attribute $a_j$. An example of such an instruction template is visualized in Figure \ref{fig:prompts}.

During the training stage, we use a differentially private optimizer to fine-tune our language model to generate each text sample within the original dataset based on the prompt corresponding to its individual attributes. Subsequently, we can synthesize a new dataset by controllably sampling text based on our desired attributes passed in the prompt. To generate a private "twin" dataset, one might use the same distribution of textual attributes as in the original dataset. Alternatively, the instruction-based approach allows us to control and change such ratios, for instance if we desire to debias our original data.

\subsection{Reducing faulty labels with prompt-mismatch objective}

The standard training objective for autoregressive language modeling is to minimize the negative log-likelihood (NLL) of every token given its previous tokens. We incorporate the natural language instructions \citep{radford2019language, NEURIPS2020_1457c0d6} into this training objective.
For every text sequence $\bm{x}$ and its corresponding attribute values $a := (a_1, ..., a_M)$, we construct the concatenated sequence  $\bm{i}(a) \oplus \bm{x}$ which prepends a corresponding task instruction to each text sample.
Let $L$ denote the length of this concatenated sequence and let $w_l$ be the sequence's $l$-th token. Our NLL loss is now
\begin{align}
    \mathrm{NLL}(\bm{i}(a) \oplus \bm{x}) = - \sum_{w_l \in \bm{i}(a) \oplus \bm{x}} \log p({w}_l|w_{<l})
    ~.
\end{align}


This objective encourages the model to generate correct samples for a given instruction. However, it does not minimize the likelihood of generating wrong samples corresponding to another prompt and therefore attribute. This is specifically unfavorable for our goal of generating synthetic training datasets as every generated text having an error of this kind corresponds to a wrongly labeled training sample. To address this, we extend the training objective with a term penalizing the generation of a given sample for a wrong prompt. Specifically, let $I_{\mathrm{wrong}}$ denote the set of all prompts not matching the given attribute values $a_1, ..., a_M$, specifically

\begin{align}
    I_{\mathrm{wrong}} := \{\bm{i}(\overline{a}_1, ..., \overline{a}_M)\text{ }|\text{ } \overline{a}_j \in C_j\setminus \{a_j\}\}~.
\end{align}

We now define the overall training loss we are aiming to minimize as
\begin{align}
\begin{split}
    &\mathcal{L}_{\mathrm{ovr}} = 
    \mathrm{NLL}(\bm{i}(a)\oplus \bm{x}) \\
    & - \frac{\lambda}{|I_{\mathrm{wrong}}|} \sum_{i_\mathrm{w}\in I_{\mathrm{wrong}}} \mathrm{NLL}(\bm{i}_\mathrm{w}\oplus {x})
    )
    ~,
\end{split}
\end{align}
where \(\lambda\) is the hyperparameter to balance the two losses. Note that in practice, when the number of possible labels is high, this computation might be inefficient and the objective too complex for the model to realize. In this case, one might randomly sample a few class labels for the wrong prompt in every training batch or penalize the generation for class labels that are the most similar to the correct one.

\begin{table*}
\centering
\small

    \caption{Accuracy of classification models trained on synthetic data.} 
    \label{tab:results}
    \setlength{\tabcolsep}{7pt} 
    \begin{tabular}{rccccccccc}
    \toprule
        & \multicolumn{3}{c}{IMDb} & \multicolumn{6}{c}{Amazon}\\
        \cmidrule(lr){2-4} \cmidrule(lr){5-10}
        & \multicolumn{3}{c}{Sentiment} & \multicolumn{3}{c}{Sentiment} & \multicolumn{3}{c}{Product Category}\\
        \cmidrule(lr){2-4} \cmidrule(lr){5-7} \cmidrule(lr){8-10}
        \# Train Samples & 25 & 50 & 5000 &25 & 50 & 3000 & 25 & 50 & 3000\\
        \midrule
        \textbf{BERT:} & \multicolumn{8}{c}{} \\
        $\epsilon$ = 3 & 82.8\% & 88.3\% & 89.1\% & 85.2\% & 87.2\% & 88.5\% & 98.6\% & 98.7\% & 98.9\% \\
        $\epsilon$ = 8 & 86.0\% & 87.6\% & 89.1\% & 87.4\% & 85.9\% & 89.2\% & 98.5\% & 98.9\% & 98.9\% \\ 
        $\epsilon$ = $\infty$ & 86.5\% & 87.6\% & 89.2\% & 89.2\% & 88.5\% & 89.2\% & 98.7\% & 98.8\% & 99.0\% \\
        \midrule
        \textbf{TF-IDF:} & \multicolumn{8}{c}{} \\
        $\epsilon$ = 3 & 71.7\% & 78.3\% & 81.0\% & 69.5\% & 75.4\% & 79.1\% & 96.8\% & 97.0\% & 98.0\%\\
        $\epsilon$ = 8 & 76.4\% & 79.2\% & 82.6\% & 74.9\% & 74.5\% & 78.3\% & 96.8\% & 98.2\% & 98.2\%\\
        $\epsilon$ = $\infty$ & 80.2\% & 79.0\% & 82.5\% & 75.2\% & 77.9\% & 79.7\% & 97.6\% & 97.9\% & 98.1\% \\
        \bottomrule
    \end{tabular}
\end{table*}

\section{Evaluation}

We conduct extensive evaluation measuring the utility and privacy of our generated data as well as the quality of its language. In this section, we describe the datasets we use as well as our evaluation metrics, implementation details and results.

\subsection{Datasets}

We use two publicly available datasets that are widely used for evaluating the performance of text classification models:

\subsubsection{IMDb Movie Reviews} The IMDb movie review dataset\footnote{\url{https://datasets.imdbws.com/}} consists of movie reviews written by various authors. We use the two binary sentiment labels as attributes to condition our model on and use a random subset of 5,000 reviews for training and evaluation each.

\subsubsection{Amazon Multi Domain Reviews} The Amazon multi domain review dataset was introduced by \citet{blitzer-etal-2007-biographies} and consists of two thousand product reviews from each of the four product categories books, DVDs, electronics and kitchen appliances. Both binarized sentiment labels and the product categories books and electronics serve as attributes we consider. Our resulting training data consists of 3,000 training samples and 1,000 test samples.

\subsection{Implementation Details}

We implement and train our language models using the PyTorch \citep{NEURIPS2019_bdbca288} and Hugging Face Transformers \citep{wolf-etal-2020-transformers} libraries and the 1.5B parameter implementation of GPT-2 \citep{radford2019language}. 
To fine-tune the language models, we employ the \enquote{privacy engine}
of the \texttt{private-transformers}\footnote{\url{https://github.com/lxuechen/private-transformers}}
package by \citet{li2021large}.
In line with their experiments, we also use DP-Adam \cite{dong2019gaussian,bu2020deep},
a differentially private version of the Adam \cite{kingma2014adam} optimizer.
The privacy engine allows us to specify desired target privacy parameters $\epsilon$ and $\delta$,
from which the standard deviation parameter $\sigma$ for the Gaussian mechanism is derived
using either Rényi DP \cite{mironov2017renyi}, the CLT \cite{dong2019gaussian},
or the FFT accountant \cite{gopi2021numerical}. Following \citet{li2021large}, we set $\delta = \frac{1}{2*|D_{train}|}$ and vary the parameter $\epsilon$ in our experiments.
To obtain reliable results for training our generative models on small subsets of the training samples, we sample three random subsets for every size and report averaged results from these three experimental runs. We trained GPT-2 over five epochs when using a differentially private optimizer and merely two epochs when using a non-private optimizer, as the latter tended to overfit quickly on the small training set. To further mitigate this, a smaller learning rate turned out to be more effective for non-private optimization: While we used a learning rate of 8e-6 with DP-Adam, we obtained the best results for non-private optimization with a learning rate of 5e-7. Lastly, we chose the hyperparameter \(\lambda := 0.2\). We generated our synthetic datasets using the original distribution of sentiment and product category attributes, which was 50 / 50 in all cases. To sample from GPT-2, we use nucleus sampling \citep{Holtzman2020The} with $p=0.8$ across all experiments.
Our results were not obtained through an extensive hyperparameter search but educated guesses over a couple of iterations to avoid large computational effort. All experiments were performed using a NVIDIA Tesla A100 GPU. With this setup, a training epoch over 1,000 text samples took approximately five minutes.



\begin{table}
\centering
\small

    \caption{Accuracy of classification models trained on real data.} 
    \label{tab:dpclassifier}
    \setlength{\tabcolsep}{7pt} 
    \begin{tabular}{rccc}
    \toprule
        & \multicolumn{1}{c}{IMDb} & \multicolumn{2}{c}{Amazon}\\
        \cmidrule(lr){2-2}\cmidrule(lr){3-4}
        & \multicolumn{1}{c}{Sentiment} & \multicolumn{1}{c}{Sentiment} & \multicolumn{1}{c}{Product}\\
        \midrule
        \textbf{BERT:} & \multicolumn{3}{c}{} \\
        $\epsilon$ = 3 & 83.6\% & 79.5\% & 95.2\% \\
        $\epsilon$ = 8 & 86.7\% & 83.4\% & 96.6\% \\ 
        $\epsilon$ = $\infty$ & 90.9\% & 91.2\% & 98.9\% \\
        \midrule
        \textbf{TF-IDF:} & \multicolumn{3}{c}{} \\
        $\epsilon$ = $\infty$ & 85.5\% & 75.8\% & 98.6\% \\        \bottomrule
    \end{tabular}
\end{table}

\subsection{Experimental Results}
As stated previously, we aim to synthesize datasets that (1) reflect properties of the original data and can be used to train classifiers that perform similar to those trained on the original data, (2) are private and do not leak information from the original data and (3) are diverse and of high language quality. Accordingly, we perform experiments with metrics maesuring these attributes and report our results in the following:

\subsubsection{Data Utility} To measure the utility of our datasets, we train classification models for each attribute on both our original data and the generated data and compare their performances when evaluating them on our real test data. Ideally, our anonymized twin datasets should lead to classifiers that are as accurate as those trained on our original data. To account for various settings including those with computational constraints, we train a shallow support vector machine classifier based on Tf-idf encodings as well as a deep BERT \citep{devlin-etal-2019-bert} based classifier with 110M parameters. Furthermore, as an interesting baseline, we evaluate the performance of BERT trained on real data with differentially private optimization using the code from \citet{li2021large}.
The classification accuracies for models trained on synthetic data are shown in Table \ref{tab:results} and classification accuracies for models trained on real data are shown in \ref{tab:dpclassifier}. In the following, we summarize our key findings:

\paragraph{Synthetic data is almost on par with real data:}

The performance of classification models trained on generated data only drops minimally compared to those trained on real data. Across all three classification tasks, for both datasets with privacy budgets of $\epsilon=3$ and $\epsilon=8$, the accuracy of BERT and Tf-Idf based models is never less than 3\% of the accuracy obtained for real data when given access to all training samples (5,000 and 3,000 for IMDb and Amazon, respectively).

\paragraph{Classifiers trained on synthetic data outperform private classifiers trained on real data:}

Notably, when comparing the results of transformer based classifiers trained on our synthetic data (Table \ref{tab:results}) to those trained on real data with differentially private optimization (Table \ref{tab:dpclassifier}), we find that the former substantially outperforms the latter across all tasks for both $\epsilon=3$ and $\epsilon=8$. This raises the question whether the intermediate step of private data generation should always be performed rather than training classifiers with DP-SGD.

\paragraph{Private data generation shows high utility in few-shot settings:}

Lastly, even when given only as little as 25 or 50 samples, GPT-2 can generate datasets that lead to high-performing classifiers, which can most likely be attributed to the utilization of pretraining knowledge through our prompting techniques. Therefore, beyond the anonymization of existing datasets, our method can be used to enlarge existing small datasets in a private manner.

\subsubsection{Data Privacy} To the best of our knowledge, methods aiming to measuring the privacy of textual data are an active area of research \citep{lm-extractdata, brown-preserve-privacy} and there is no standardized and agreed upon way to do so.
In our experiments, we follow \citet{lm-extractdata} by counting the number of instances in which our synthetic dataset contains samples that are extremely close to a sample from the training data and can therefore be considered a duplicate: For every sample $\bm{x}_i$ from our training data used for the language model and every $\bm{x}_j \in \widetilde{D}_{\mathrm{train}}$, we measure the set of trigrams $\mathrm{g_3}(\bm{x}_i)$, $\mathrm{g_3}(\bm{x}_j)$. We consider the two samples as duplicates if \[|\mathrm{g_3}(\bm{x}_i) \cup \mathrm{g_3}(\bm{x}_j)| \geq 2*\mathrm{min}(|\mathrm{g_3}(\bm{x}_i)|, |\mathrm{g_3}(\bm{x}_j)|)\]
As we hypothesize that duplicates are relatively rare, we double the generated data compared to our utility experiments and search for them within 10,000 and 6,000 samples generated for the IMDb and Amazon dataset, respectively. 
Our results are depicted in Table \ref{tab:privacyresults} and demonstrate the significant reduction of data leakage from privately trained models. 

\begin{table}[h]
  \centering
  \small
\setlength\tabcolsep{7pt}
  \begin{tabular}{ r c c c c c c}
    \toprule
    & \multicolumn{3}{c}{IMDb}& \multicolumn{3}{c}{Amazon}\\
    \cmidrule(lr){2-4} \cmidrule(lr){5-7}
    \# Samples & 25 &  50 & 5000 & 25 &  50 & 3000\\
    \midrule
    $\epsilon = 3$ & 1 & 0 & 1 & 0 & 0 & 0\\
    $\epsilon = 8$ & 0 & 6 & 2 & 4 & 1 & 0\\
    $\epsilon = \infty$ & 8 & 23 & 13 & 16 & 30 & 9\\
    \bottomrule

  \end{tabular}
  \caption{Number of duplicates from the training data generated by language models}
  \label{tab:privacyresults}
\end{table}

\subsubsection{Language Quality} As a metric measuring the quality of our generated samples, we use the Mauve\footnote{\url{https://github.com/krishnap25/mauve}} \citep{pillutla2021mauve} score to compute the similarity of the distributions of $\mathcal{D}_{\mathrm{train}}$ and the generated data $\widetilde{\mathcal{D}}_{\mathrm{train}}$ from every trained model. As can be seen in Table \ref{tab:qualityresults}, higher \(\epsilon\) values tend to increase the quality measured by Mauve, but overall seem not to be highly significant. As a reference, the Mauve score computed when comparing \(\mathcal{D}_{train}\) and \(\mathcal{D}_{test}\) are 0.95 for IMDb and 0.94 for Amazon.
Based on manual inspection, the quality of generated texts seems to be very high. Mismatches between prompts and generated texts (e.g. a negative review generated for a positive prompt) as well as incoherent generations do occur, but very rarely. Excerpts of the generated data can be seen in Table \ref{tab:examples}, failure cases can be found in Table \ref{tab:failurecases} and \ref{tab:failurecaseslanguage} in the appendix.

\begin{table}[h]
  \centering
  \small
\setlength\tabcolsep{5pt}
  \begin{tabular}{ r c c c c c c}
    \toprule
    & \multicolumn{3}{c}{IMDb}& \multicolumn{3}{c}{Amazon}\\
    \cmidrule(lr){2-4} \cmidrule(lr){5-7}
    \# Samples & 25 &  50 & 5000 & 25 &  50 & 3000\\
    \midrule
    $\epsilon = 3$ & 0.81 & 0.83 & 0.81 & 0.82 & 0.81 & 0.83\\
    $\epsilon = 8$ & 0.82 & 0.81 & 0.81 & 0.82 & 0.81 & 0.82\\
    $\epsilon = \infty$ & 0.81 & 0.85 & 0.84 & 0.84 & 0.85 & 0.82\\
    \bottomrule

  \end{tabular}
  \caption{Mauve scores measuring the similarity of generated data and \(\mathcal{D}_{train}\).}
  \label{tab:qualityresults}
\end{table}

\begin{table*}
\centering
\small

    \caption{Exemplary generated samples from models trained with only 25 text samples. The texts were selected by picking a random sample from \(\mathcal{D}_{test}\) and finding the most similar one according to Sentence-BERT \citep{reimers-gurevych-2019-sentence} within the generated data.} 
    \label{tab:examples}
    \setlength{\tabcolsep}{7pt} 
    \begin{tabular}{l l}
    \toprule
        \multicolumn{2}{c}{\textbf{Amazon (Category=Electronics, Sentiment=Positive)}}\\
        \midrule\\
        Original & I'm not any expert in the computer world. But this product it is very easy to install. The range is very strong.\\ & I'm very happy with this product!!!\\\\
        $\epsilon = 3$ & I really like this. It works great. The speed is a bit slow. But, after I learned that this is how a good \\ & keyboard feels. I can't say enough good things about this keyboard.\\\\
        $\epsilon = 8$ & This is a good piece of equipment. It is small, has a high quality and a reasonable price. I used it to test the\\&sound from my computer monitor. I am happy with the sound.\\\\
        $\epsilon = \infty$ & I got this game to play on my PSP as a little side project. I'm quite happy with it.   The graphics are pretty \\&good and the story line is pretty entertaining, as well.\\\\
        \midrule
        \multicolumn{2}{c}{\textbf{Amazon (Category=Books, Sentiment=Negative)}}\\
                \midrule\\
        Original & Good Lord, people must be busier than I thought. If you haven't been allowed to even think about your future\\& after retirement, then maybe you will get something out of this book.  For the rest of us, this is drivel\\\\
        $\epsilon = 3$ & I have been looking for an awesome, fun, easy to read book, like a classic from when I was a kid. I am very \\&disappointed. It is not a fun read. It really is a bore. I have read the sequel, and am also disappointed.\\& This is not one of those things that you read to pass the time. Not a good choice for a first book.\\\\
        $\epsilon = 8$ & I would've done better with another book I read. The plot is interesting but the characters are bland and\\& the setting is really only a backdrop to the action and plot. It's a shame as I enjoyed the book, but\\& this isn't a good read.  \\\\
        $\epsilon = \infty$ & This book was boring, boring, and boring. I have been thinking about getting a new copy of this book ever \\&since I read it, but this one didn't work for me at all. Not a bad idea, just not my cup of tea.\\
        \bottomrule

    \end{tabular}
\end{table*}

\section{Related Work}

\subsection{Text Anonymization} 
Substantial efforts have been made to enable the privacy-preserving processing of textual data through both private textual vector representations and by transforming text into readable anonymous formats. Approaches from the former category either aim at obtaining term frequency vectors using differentially private mechanisms \cite{syntf-weggenmann, fernandes2019generalised} or by using deep learning methods with adversarial training objectives \citep{coavoux-etal-2018-privacy}. In the work by \citet{nlu-private-bert}, various local DP mechanisms are explored to obtain private BERT representations.

Methods aiming at rewriting texts in a privacy-preserving manner range from rule-based approaches using human-engineered text perturbations \citep{mahmood-mutantx, bevendorff-etal-2019-heuristic} as well as word replacements through the perturbation of individual word embeddings using differential privacy \citep{feyisetan2019leveraging, feyisetan2020privacy} to deep learning based approaches leveraging sequence-to-sequence models. These sequence-to-sequence models can either incorporate adversarial objectives penalizing the generation of author-revealing information \citep{a4nt-shetty, xu-etal-2019-privacy} or integrate differential privacy in the text sampling process \citep{bo-etal-2021-er, dp-vae-weggenmann, mattern-limits}.

Notably, various papers proposing the integration of differentially private mechanisms in deep learning architectures \citep{krishna-etal-2021-adept, dp-text1, dptext2, dptext3} have been shown to actually violate differential privacy \citep{habernal2021differential, habernal-reparametrization}. While these works still represent important contributions due to their good empirical results, it should be noted that the design of NLP systems with DP guarantees is a task that is prone to errors and should be approached carefully.

\subsection{Differentially Private Language Model Training} 
As generative language models have been shown to leak training data \citep{lm-extractdata} and the embeddings of discriminative models have been shown to contain sensitive information about a text's originator \citep{embedding-info-leakage},
differentially private optimizers such as DP-SGD \citep{dp-sgd, bassily-private-erm} and DP-Adam \citep{deeplearning-dp, adam-optimizer} have been applied to a variety of NLP tasks. Large-scale pretraining of BERT using DP-SGD has shown to reap comparable masked language modeling performance to non-private BERT \citep{large-scale-dp-bert}. For the tasks of text classification and named entity recognition, good performance has been obtained with BERT and DP-SGD, but only with large privacy budgets of \(\epsilon = 100\) or higher. Recently, it has been demonstrated that with the correct choice of hyperparameters and fine-tuning objectives aligned with the pretraining procedure, both generative and discriminative language models can achieve high performance in various tasks even with stricter privacy bounds \citep{li2021large, yu2022differentially}. An active area of research is concerned with the empirical evaluation of a language model's privacy  \citep{brown-preserve-privacy} using methods such as membership inference attacks \citep{hayes2019logan}.

\subsection{Synthetic Data Generation:}
Synthetic data generation with privacy guarantees using methods such as DP-GAN \citep{dp-gan}, PATE-GAN \citep{jordon2018pate} or various related approaches has successfully been applied for structured tabular or visual data \citep{Torkzadehmahani_2019_CVPR_Workshops, neunhoeffer2020private, chan-gs-wgan}. Beyond these methods, DPRP (Differentially Private Data Release via Random Projections) \citep{pmlr-v124-gondara20a} has been proposed as a model free alternative for releasing small private datasets that does not require training a generative model. For the domain of text, synthetic data generation techniques have predominantly been developed and evaluated without considering privacy guarantees \citep{Anaby-lambada, schick2021generating}. Merely the work presented by \citet{bommasani2019towards} is similar to our paper, but does not provide any quantitative results about the experiments.

\section{Conclusion}

In this paper, we explored the generation of synthetic datasets from differentially private language models as a solution for publicly sharing textual data while protecting the privacy of users whose data is being shared. Our experiments show that synthetic data from differentially private language models is of high quality and is very well suited as training data for further tasks while significantly reducing the risk of leaking the original data. Our approach can be applied in a variety of use cases working with sensitive data. An interesting challenge for future work is the anonymization of multimodal datasets consisting of tabular, visual and text data.

\section*{Limitations}
\paragraph{Privacy Guarantee} While differential privacy provides a statistical privacy guarantee, one can not be certain that a differentially private language model does not leak any sensitive information. As seen in our experiments, the differentially private models did leak some of their training data, even if significantly less than the non-private ones. This can be a concern when dealing with training data containing names, telephone numbers or even passwords.

\paragraph{Synthetic Data Quality} As shown in Table \ref{tab:failurecases} and  \ref{tab:failurecaseslanguage}, our models did in rare cases produce incoherent language or text samples that did not fit the desired control attributes. This can limit the quality of the generated data.

\paragraph{Limits of Controllable Generation} The controllability of multiple fine-grained textual attributes in text generation remains a difficult challenge \citet{lyu-etal-2021-styleptb}. We therefore need to assume that our approach will become less accurate the higher the amount of textual attributes we want to consider.

\section*{Ethical Considerations}
Data privacy is a highly important issue for the responsible deployment of machine learning solutions. With our work, we directly contribute to this field of research. 

As our method relies on large pretrained language models, it should be noted that users deploying these technologies need to be aware of their undesirable, human-like biases \citep{sheng-etal-2019-woman, llm-muslim}. Methods for reducing these harmful associations are actively being developed by the research community \citep{liang2021towards, schick-diagnosis}.

\section*{Acknowledgments}
This material is based in part upon works supported
by the German Federal Ministry of Education and Research (BMBF): Tübingen AI Center, FKZ: 01IS18039B;
by the German Federal Ministry for Economic Affairs and Climate Action (BMWK) in the project \emph{Trade-EVs II}, FKZ: 01MV20006A;
by the Machine Learning Cluster of Excellence, EXC number 2064/1 – Project number 390727645; by the John Templeton Foundation (grant \#61156); by a Responsible AI grant by the Haslerstiftung; and an ETH Grant
(ETH-19 21-1).
Zhijing Jin is supported by PhD fellowships from the Future of Life Institute and Open Philanthropy, as well as the travel support from ELISE (GA no 951847) for the ELLIS program. We also thank OpenAI Researcher Access Program for granting our team credits to their API.

\bibliography{anthology,custom}
\bibliographystyle{acl_natbib}

\appendix

\section{Failure Cases}
We demonstrate manually selected failure cases in Tables \ref{tab:failurecases} and \ref{tab:failurecaseslanguage}.
Generated texts from Table \ref{tab:failurecases} do not reflect the attributes passed in the prompt accurately and would therefore correspond to wrongly labeled samples in our dataset. Table \ref{tab:failurecaseslanguage} features generated texts that contain logical or language errors.

\begin{table*}[t]
\centering
\small

    \caption{Failure cases in which the generated text does not fit the desired attributes. Model mistakes are marked in \color{red}red.}
    \label{tab:failurecases}
    \setlength{\tabcolsep}{5pt} 
    \begin{tabular}{c c c l}
    \toprule
    \# Train Samples & $\epsilon$ & Desired Attributes & Text\\
    \midrule\\{\multirow{7}{*}{\shortstack[c]{25}}}& {\multirow{7}{*}{\shortstack[c]{3}}}& {\multirow{7}{*}{\shortstack[c]{Category=Books\\{\color{red}Sentiment = Negative}}}}&
      {\multirow{7}{*}{\shortstack[l]{I have read and enjoyed many of your books, but I can't recommend this\\ one highly enough. I am not a big fan of vampires. They don't interest me \\at all This book, however, did. It was a fun, entertaining read. I loved the\\characters. I really didn't care if I was going to be able to go back in time to\\help save the future, or if I would be in for the long haul. There was a lot of\\tension throughout the book. I was glad I was able to finish it. I'm glad that\\I was not as upset with it as many others were. I gave this book a solid 5 stars.}}} \\\\\\\\\\\\\\\\
      
      \midrule\\{\multirow{3}{*}{\shortstack[c]{50}}}& {\multirow{3}{*}{\shortstack[c]{3}}}& {\multirow{3}{*}{\shortstack[c]{Category=Books\\{\color{red}Sentiment = Negative}}}}&
      {\multirow{3}{*}{\shortstack[l]{I was looking for a book on how to be a good leader and found it by chance.\\ It seems that no matter how bad you have been, how bad things have gotten,\\or how terrible a person you are, a leader can still turn it around.}}} \\\\\\\\
      
      \midrule\\{\multirow{4}{*}{\shortstack[c]{50}}}& {\multirow{4}{*}{\shortstack[c]{3}}}& {\multirow{4}{*}{\shortstack[c]{{\color{red}Category=Books}\\Sentiment = Positive}}}&
      {\multirow{4}{*}{\shortstack[l]{I am not a huge fan of the Star Wars films but I really enjoyed this one.\\ It is the first time I've ever seen the Star Trek movies and I enjoyed\\ them as well.  The plot is a bit different then you'd expect from a Star Trek\\story but I enjoyed it all the same.}}} \\\\\\\\\\
      
      \midrule\\{\multirow{5}{*}{\shortstack[c]{50}}}& {\multirow{5}{*}{\shortstack[c]{8}}}& {\multirow{5}{*}{\shortstack[c]{Category=Electronics\\{\color{red}Sentiment = Negative}}}}&
      {\multirow{5}{*}{\shortstack[l]{I have a brand new LG Electronics Vivid 5, and so far I'm very happy.\\  The screen has a nice clear picture and has great picture quality, the\\ speakers are good, the microphone is not great, the battery is adequate,\\ and it's nice and light.    This is the first of the Vivid series I have bought, and\\ I'm glad I got one.}}} \\\\\\\\\\\\

\bottomrule

    \end{tabular}
\end{table*}

\begin{table*}[t]
\centering
\small

    \caption{Failure cases in which the generated text is incoherent or does not make sense. Model mistakes are marked in \color{red}red.}
    \label{tab:failurecaseslanguage}
    \setlength{\tabcolsep}{5pt} 
    \begin{tabular}{c c c l}
    \toprule
    \# Train Samples & $\epsilon$ & Desired Attributes & Text\\
    \midrule\\{\multirow{7}{*}{\shortstack[c]{50}}}& {\multirow{5}{*}{\shortstack[c]{$\infty$}}}& {\multirow{5}{*}{\shortstack[c]{Category=Electronics\\Sentiment = Negative}}}&
      {\multirow{5}{*}{\shortstack[l]{The device itself, {\color{red}while very good, is bad}. It seems like the software doesn't\\ like to write out its own ID (it should) so the app doesn't work. I've seen\\other users say the same thing and have had no problems. I haven't bought another\\ bluetooth device, but I'm pretty sure it has something to do with it.}}} \\\\\\\\\\\\
      
      \midrule\\{\multirow{4}{*}{\shortstack[c]{25}}}& {\multirow{4}{*}{\shortstack[c]{8}}}& {\multirow{4}{*}{\shortstack[c]{Category=Books\\Sentiment = Positive}}}&
      {\multirow{4}{*}{\shortstack[l]{I really enjoyed this book. I didn't get the full explanation as I am only\\ a beginner but for beginners with the basic knowledge of a math and physics\\ the theory is very sound.    {\color{red}2.0 stars.}}}} \\\\\\\\\\
      \midrule\\{\multirow{4}{*}{\shortstack[c]{4,000}}}& {\multirow{4}{*}{\shortstack[c]{$\infty$}}}& {\multirow{4}{*}{\shortstack[c]{Category=Books\\Sentiment = Negative}}}&
      {\multirow{4}{*}{\shortstack[l]{{\color{red}This book is a compilation of some of my personal thoughts and ideas.}\\{\color{red}  The author's goal was to make me believe in witchcraft, but he}\\{\color{red} completely failed.}}}} \\\\\\\\\\
      
      \midrule\\{\multirow{4}{*}{\shortstack[c]{50}}}& {\multirow{4}{*}{\shortstack[c]{8}}}& {\multirow{4}{*}{\shortstack[c]{Category=Electronics\\Sentiment = Positive}}}&
      {\multirow{4}{*}{\shortstack[l]{This is the best little portable   guitar i've   bought   since the little portable\\ juke box   on a   bike.   The   high {\color{red}quality   quality  quality} of   this   product   is   great  .\\   I do   wish   {\color{red}i had bought  a  sooner   the   old   jukebox}   but   once   you\\   get   it   together   its   really   nice   to have.}}} \\\\\\\\\\

\bottomrule

    \end{tabular}
\end{table*}

\end{document}